\documentclass[11pt,a4paper]{article}
\usepackage[nohyperref]{naaclhlt2018}
\usepackage[]{naaclhlt2018}
\usepackage{times}
\usepackage{latexsym}

\usepackage{url}
\usepackage{adjustbox}
\usepackage{graphicx}
\usepackage{booktabs}
\usepackage{diagbox}
\usepackage{multirow}
\usepackage[T1]{fontenc}
\usepackage{color,soul}
\usepackage[table]{xcolor}
\usepackage{colortbl}
\usepackage{amssymb}
\usepackage{pifont}
\usepackage{xspace}
\newcommand{\cmark}{\ding{51}}%
\newcommand{\xmark}{\ding{55}}%
\usepackage{caption}
\usepackage{amsmath}
\usepackage[spanish]{babel}
\usepackage[utf8]{inputenc}
\usepackage{lingmacros}

\usepackage{lipsum}

\usepackage[disable]{todonotes} 
\usepackage{array}
\newcolumntype{h}{>{\setbox0=\hbox\bgroup}c<{\egroup}@{}}

\newcommand{\tVect}{\mathbf{u}} 
\newcommand{\hVect}{\mathbf{v}}

\newcommand{\context}{context}
\newcommand{\hypothesis}{hypothesis}
\newcommand{\reals}{\mathbb{R}}

\aclfinalcopy 
\def\aclpaperid{1041} 

\newcommand{\figref}[1]{Figure~\ref{#1}}
\newcommand{\tabref}[1]{Table~\ref{#1}}

\newcommand{\secref}[1]{\S\ref{#1}}
\newcommand{\appref}[1]{Appendix~\ref{#1}}

\newcommand\capsize{\capsizeinner}

\title{On the Evaluation of Semantic Phenomena in\\Neural Machine Translation Using Natural Language Inference}

\author{Adam Poliak$^{1}$ \hspace{.5cm} Yonatan Belinkov$^{2}$ \hspace{.5cm} James Glass$^{2}$ \hspace{.5cm} Benjamin Van Durme$^{1}$\\
\hspace{0cm} $^{1}$Center for Language and Speech Processing \\ 
\hspace{0cm} Johns Hopkins University, Baltimore, MD 21218 \\
\hspace{0cm} $^{2}$Computer Science and Artificial Intelligence Laboratory \\
\hspace{0cm} Massachusetts Institute of Technology, Cambridge, MA 02139 \\ 
\hspace{0cm} {\normalsize \tt \{azpoliak,vandurme\}@cs.jhu.edu}, {\normalsize \tt \{belinkov,glass\}@mit.edu}}

\date{}
\begin{document}
\maketitle
\begin{abstract}
We propose a process for investigating the extent to which sentence
representations arising from neural machine translation (NMT) systems
encode distinct semantic phenomena. 
We use 
these representations as features to train a natural language inference (NLI) 
classifier based on datasets recast from existing semantic annotations.
In applying this process to a representative NMT system, we find
its encoder appears most suited to supporting inferences at the
syntax-semantics interface, as compared to 
anaphora resolution requiring world-knowledge.
We conclude with a discussion on the merits and potential deficiencies
of the existing process, and how it may be improved and extended as a
broader framework for evaluating semantic coverage.\footnote{Code 
developed 
and data used are
available at \url{https://github.com/boknilev/nmt-repr-analysis}.}
\end{abstract}

\section{Introduction}
What do neural machine translation (NMT) models learn about semantics? Many researchers suggest that state-of-the-art NMT models learn 
representations that capture the meaning of sentences~\cite{gu-EtAl:2016:P16-1,TACL1081,P17-2092,D17-1311,neubig2017neural,koehn2017neural}. 
However, there is limited understanding of how specific semantic phenomena are captured in NMT representations beyond this broad notion. For instance,  
how well do these representations capture \newcite{dowty1991thematic}'s thematic proto-roles? 
Are these representations sufficient for understanding paraphrastic inference? Do the sentence representations encompass complex anaphora resolution?  
We argue that existing semantic annotations recast as Natural Language Inference (NLI)
can 
be leveraged to investigate 
whether sentence representations encoded by NMT models 
capture these 
semantic phenomena.

We use sentence representations from 
pre-trained 
NMT encoders as features to train classifiers 
for 
NLI, 
the task of
determining if one sentence (a \textit{hypothesis}) is supported by another (a \textit{context}).\footnote{Sometimes referred to as recognizing textual entailment~\cite{dagan2006pascal,dagan2013recognizing}.}
If the sentence representations learned by NMT models capture distinct semantic phenomena, we hypothesize that those representations should be sufficient to perform 
well on 
NLI datasets that  
test a model's ability to capture these phenomena.
\figref{fig:rte} shows example 
NLI sentence pairs with their respective labels and semantic phenomena.

\begin{figure}[t!]
 \centering
 \small
 \renewcommand{\arraystretch}{1.35}
 \begin{tabular}{c|c|c}
     \toprule 
  &\cellcolor{gray!20}Sara adopted Jill, \textit{she} wanted a child&\\
 \multirow{-2}{*}{DPR}&Sara adopted Jill, \textit{Jill} wanted a child& \multirow{-2}{*}{\xmark}\\ \midrule 
 &\cellcolor{gray!20}Iran \textit{possesses} five research reactors & \\
 \multirow{-2}{*}{FN+}&Iran \textit{has} five research reactors &\multirow{-2}{*}{\cmark}\\ \midrule 
 &\cellcolor{gray!20}Berry Rejoins WPP Group & \\
 \multirow{-2}{*}{SPR}&Berry was \textit{sentient}& \multirow{-2}{*}{\cmark}\\ \bottomrule 
 \end{tabular}
 \caption{\capsize Example sentence pairs for the different semantic phenomena. DPR deals with complex anaphora resolution, FN+ is concerned with paraphrastic inference, and SPR covers \newcite{TACL674}'s semantic proto-roles. 
 \cmark \space / \xmark \space  indicates that the first sentence  entails / does not entail the second.
}
 \label{fig:rte}
 \end{figure}

We evaluate  NMT sentence representations of 4 NMT models from 2 domains 
on $4$
different NLI 
datasets to investigate how well they 
capture different semantic phenomena. We use \newcite{white-EtAl:2017:I17-1}'s \textit{Unified Semantic Evaluation Framework} (USEF) 
that recasts three semantic phenomena NLI:
1) semantic proto-roles, 2) paraphrastic inference, 3) and complex
anaphora resolution.
Additionally, we evaluate the NMT sentence representations
on 4) Multi-NLI, a recent extension of the Stanford Natural Language Inference dataset (SNLI)~\cite{snli:emnlp2015} that includes multiple genres and domains~\cite{williams2017broad}.
We contextualize our results with a standard neural encoder described in \newcite{snli:emnlp2015} and used in \newcite{white-EtAl:2017:I17-1}.

Based on the recast NLI datasets, our investigation suggests
that NMT encoders might learn 
more about semantic proto-roles than anaphora resolution or paraphrastic inference.
We note that
the target-side language affects how 
an NMT source-side encoder captures 
these semantic phenomena.

\section{Motivation}
\label{sec:motivation}

\paragraph{Why use recast NLI?} 
We focus on 
NLI, as opposed to a wide range of NLP taks, as a unified framework that can capture a variety of semantic phenomena 
based on arguments by \newcite{white-EtAl:2017:I17-1}. Their recast dataset enables us to study whether NMT encoders capture ``distinct types of semantic reasoning'' under just one task. 
We choose 
these specific semantic phenomena for two reasons. First, 
a long term goal is to understand how combinations of different corpora and neural architectures can contribute to a system's ability to perform general language understanding.  As humans can understand (annotate consistently) the sentence pairs used in our experiments, we would similarly like our final system to have this same capability.  
We posit that it is necessary but not necessarily sufficient for a language understanding system to be able to capture the semantic phenomena considered here. 
Second, 
we believe these semantic phenomena might be relevant for translation. 
We demonstrate this with a few examples. 

\paragraph{Anaphora}
Anaphora resolution 
connects tokens, typically pronouns, to their referents. 
Anaphora resolution should occur when translating from morphologically poor languages into some morphologically rich languages. For example, when translating ``The parent fed the child because she was hungry,'' a Spanish translation should describe \textit{the child} as \textit{la niña (fem.)} and not \textit{el niño (masc.)} since \textit{she} refers to \textit{the child}. 
Because world knowledge is often required to perform anaphora resolution~\cite{rahman-ng:2012:EMNLP-CoNLL,javadpour2013resolving}, this may enable 
evaluating whether an NMT encoder learns world knowledge. 
In this example, \textit{she} refers to \textit{the child} and not \textit{the parent} since world knowledge dictates that parents often feed children when children are hungry.

\paragraph{Proto-roles}
\newcite{dowty1991thematic}'s proto-roles may be expressed differently in different languages, and so correctly identifying them can be important for translation.
For example, English does not usually explicitly mark  {\it volition}, 
a proto-role,
except by using adverbs like {\it intentionally} or {\it accidentally}. Other languages mark volitionality by using special affixes (e.g., Tibetan and Sesotho, a Bantu language), case marking (Hindi, Sinhalese), or auxiliaries (Japanese).\footnote{
For
references and
examples, see:  \url{en.wikipedia.org/wiki/Volition_(linguistics)}. 
}  
Correctly generating these markers may require the MT system to encode volitionality on the source side. 

\paragraph{Paraphrases}
\newcite{callisonburch:2007:thesis} discusses how paraphrases help statistical MT (SMT) when 
alignments from source words to target-language words are unknown. If the alignment model can map a paraphrase of the source word to a word in the target language, then the SMT model can translate the original word based on its paraphrase.\footnote{Using paraphrases can help NMT models generate text in the target language in some settings~\cite{sekizawa-kajiwara-komachi:2017:WAT2017}.} 
Paraphrases are also used by professional translators 
to deal
with non-equivalence of words in the source and target languages~\cite{baker2018other}. 

\begin{table*}
\begin{adjustbox}{width=1\textwidth}

	\centering
\begin{tabular}[t!]{c|ccccc|ccccc|ccccc}
\toprule 
\multirow{2}{*}{\backslashbox{\small{Train}}{\small{Test}}}	 & \multicolumn{5}{c|}{DPR: 50.0}  & \multicolumn{5}{c|}{SPR: 65.4} & \multicolumn{5}{c}{FN+: 57.5} \\
 	& ar & es & zh & de & USEF &  ar & es & zh & de  & USEF &  ar & es & zh & de  & USEF \\  \midrule 
 DPR   &  49.8  & {\bf 50.0}  &  {\bf 50.0} & {\bf 50.0} & 49.5 & 45.4 & 57.1 & 47.0 & 43.9 & {\bf 65.2}& 48.0  & {\bf 55.9} & 51.0 & 46.8 & 19.2 \\ 
 SPR  & 50.1 & 50.3 & 50.1 & 49.9 & {\bf 50.7} & 72.1 & 74.2 & 73.6 & 73.1 & {\bf 80.6} & 56.3 & 57.0 & 56.9 & 56.1 & {\bf 65.8}\\ 
 FN+   & 50.0 & 50.0 & {\bf 50.4}  & 50.0 & 49.5 & 57.3 & {\bf 63.6} & 54.5 & 60.7  & 60.0 & 56.2 & 56.1 & 54.3 & 55.5 & {\bf 80.5} \\ \bottomrule 
\end{tabular}
\end{adjustbox}
\caption{\capsize Accuracy on NLI  
with representations generated by encoders of English$\rightarrow$\{ar,es,zh,de\} NMT models. 
Rows correspond to the training and validation sets and major columns correspond to the test set. 
The column labeled ``USEF'' refers to the test accuracies reported in \newcite{white-EtAl:2017:I17-1}. The numbers on the top row represents each dataset's majority baseline. Bold numbers indicate the highest performing model for the given dataset.} 
\label{tab:rte-results}
\end{table*}

\section{Methodology} 
We use  NMT models based on bidirectional long short-term memory (Bi-LSTM) encoder-decoders 
with attention~\cite{sutskever2014sequence,bahdanau2014neural}, trained on a parallel corpus. 
Given an NLI \context-\hypothesis~pair, we pass each sentence independently through a trained NMT encoder to extract their respective vector representations. 
We represent each sentence by concatenating the last hidden state from the forward and backward encoders,
resulting in $\hVect$ and $\tVect$ (in $ \reals^{2d}$) for the \context~and \hypothesis.\footnote{We
experimented with other sentence representations and their combinations, 
and 
did not see 
differences in overall conclusions. 
See \appref{app:repr} for these experiments.
}
We follow the common practice of feeding the concatenation $(\hVect, \tVect) \in \reals^{4d}$ to a classifier~\cite{rocktaschel2015reasoning,snli:emnlp2015,mou-EtAl:2016:P16-2,liu2016learning,cheng-dong-lapata:2016:EMNLP2016,munkhdalai-yu:2017:EACLlong1}.

Sentence pair representations are fed into a classifier with a softmax layer that maps onto the number of labels. 
Experiments with both linear and non-linear classifiers have not shown major differences, so we report results with the linear classifier unless noted otherwise. 
We report implementation details
in~\appref{app:exp-details}.

\section{Data}
\paragraph{MT data}
We train 
NMT models 
on 
four language pairs: English $\rightarrow$ 
\{Arabic (ar), Spanish (es), Chinese (zh), and German (de)\}.
See~\appref{app:exp-details} for training details.
The first three 
pairs use  
the United Nations parallel
corpus~\cite{ZIEMSKI16.1195} 
and for English-German, we 
use the WMT dataset~\cite{bojar-EtAl:2014:W14-33}. 
Although the entailment classifier only uses representations extracted from the English-side encoders as features, using multiple language pairs allows us to explore
whether
different 
target languages 
affect what semantic phenomena are captured by an NMT encoder.

\paragraph{Natural Language Inference data}
We use  
four distinct datasets to train classifiers: 
Multi-NLI~\cite{williams2017broad}, a recent expansion of SNLI containing a broad array of domains that was used in the 2017 RepEval shared task~\cite{nangia2017repeval},
and three recast NLI datasets from The JHU Decompositional Semantics Initiative (\texttt{Decomp})\footnote{\url{decomp.net}} 
released by
\newcite{white-EtAl:2017:I17-1}. 
Sentence-pairs and labels 
were recast, i.e. automatically converted, from existing semantic annotations: FrameNet Plus (FN+)~\cite{pavlick-EtAl:2015:ACL-IJCNLP2},
Definite Pronoun Resolution (DPR)~\cite{rahman-ng:2012:EMNLP-CoNLL}, and Semantic Proto-Roles (SPR)~\cite{TACL674}.
The FN+ portion contains sentence pairs based on paraphrastic inference,
DPR's sentence pairs focus on identifying the correct antecedent
for a definite pronoun, and  
SPR's sentence pairs test whether the semantic proto-roles from \newcite{TACL674} apply based on a given sentence.\footnote{We refer the reader to \newcite{white-EtAl:2017:I17-1} for detailed discussion on how the existing datasets 
were recast as NLI.} 
Recasting makes it easy to determine how well an NLI method captures the fine-grained semantics inspired by \newcite{dowty1991thematic}'s thematic proto-roles,  paraphrastic inference, and complex anaphora resolutions.
\tabref{tab:rte-datasets} includes 
the datasets' statistics.

\begin{table}
	\centering
\begin{tabular}[t]{l|rrr|r}
\toprule
  & DPR & SPR & FN+ & MNLI \\  
 \midrule
 Train   & 2K   & 123K & 124K & 393K \\ 
 Dev  & .4K & 15K & 15K & 9K \\ 
 Test   & 1K & 15K & 14K & 9K \\ 
 \bottomrule
\end{tabular}
\caption{\capsize
Number of sentences in NLI datasets.
}
\label{tab:rte-datasets}
\end{table}

\section{Results} \label{sec:analysis} 
\tabref{tab:rte-results} shows 
results of NLI classifiers trained on 
representations from different NMT encoders. 
We also report the majority baseline and the results of \citeauthor{snli:emnlp2015}'s  3-layer deep 
200 dimensional neural network 
used by \citeauthor{white-EtAl:2017:I17-1} 
(``USEF'').

\paragraph{Paraphrastic entailment (FN+)}
Our classifiers predict FN+ entailment worse than the  majority baseline, and drastically worse than USEF when trained on FN+'s training set. Since FN+ tests paraphrastic inference and NMT models have been shown to be useful to generate sentential paraphrase pairs~\cite{wieting2017pushing,wieting-mallinson-gimpel:2017:EMNLP2017}, it is surprising that our classifiers using the representations from the NMT encoder perform poorly. 
Although the sentences in FN+ are much longer than in the other datasets,
sentence length does not seem to be responsible for the poor FN+ results. The classifiers do  
not noticeably perform better on shorter sentences than longer ones, as noted in~\appref{app:data-diff}.

Upon manual inspection, we noticed that in many \textit{not-entailed} examples, swapped paraphrases had different part-of-speech (POS) tags. This begs the question of whether different POS tags for swapped paraphrases 
affects the accuracies. 
Using Stanford CoreNLP~\cite{manning-EtAl:2014:P14-5}, we partition our validation set based on whether the paraphrases share the same POS tag. ~\tabref{tab:fnplus-pos-results} reports dev set accuracies  
using classifiers 
trained on FN+. 
Classifiers using features from 
NMT encoders trained on the three languages from the UN corpus noticeably perform better on cases where paraphrases have different POS tags compared to paraphrases with the same POS tags. These differences might suggest that the recast FN+ might not be an ideal dataset to test how well NMT encoders 
capture paraphrastic inference. The sentence representations may be impacted more by ungrammaticality caused by different POS tags as opposed to poor paraphrases.
\begin{table}[t]
	\centering
    \small
	\begin{tabular}{l|cccc} 
	  	\toprule 
		& ar & es & zh & de  \\ \midrule 
 		Same Tag & 52.9 & 52.6 & 52.6 & 50.2 \\ 
		Different Tag & 55.8 & 59.1 & 53.4 & 46.0 \\ 
		\bottomrule 
    \end{tabular}
    \caption{\capsize Accuracies on FN+'s dev set based on whether the swapped paraphrases share the same POS tag. 
    }
    \label{tab:fnplus-pos-results}
\end{table}

\paragraph{Anaphora entailment (DPR) }
The low accuracies for predicting NLI targeting 
anaphora resolution 
are similar to \newcite{white-EtAl:2017:I17-1}'s findings. They suggest 
that 
the model has difficulty in capturing complex anaphora resolution. 
By using contrastive evaluation pairs, \newcite{bawden2017evaluating} recently suggested as well that NMT models are poorly suited for co-reference resolution. 
Our results are not surprising given that DPR tests whether a model contains common sense knowledge~\cite{rahman-ng:2012:EMNLP-CoNLL}. 
In DPR, syntactic cues for co-reference are purposefully balanced out as each pair of pro-nouns appears in at least two context-hypothesis pairs 
(\tabref{tab:dpr-example}). This forces the model's decision to be 
informed by semantics and world knowledge --
a model 
cannot use syntactic cues to help perform anaphora resolution.\footnote{\appref{sec:wk-dpr} includes some illustrative examples.} 
Although the poor performance of NMT representations may be explained by 
a variety of reasons, e.g. training data, architectures, etc., we would still like ideal MT systems to capture the semantics of 
co-reference, as evidenced in the example in \secref{sec:motivation}. 

Even though the classifiers perform poorly when predicting 
paraphrastic entailment, 
they surprisingly outperform USEF 
by a large margin (around $25$--$30\%$) 
when using a model trained on 
DPR.\footnote{This is seen in the last columns of the top row in~\tabref{tab:rte-results}.}
This might suggest that 
an NMT 
encoder can pick up on how pronouns 
may be 
used as 
a type of lexical paraphrase~\cite{bhagat2013paraphrase}.

\paragraph{Proto-role entailment (SPR)}
When predicting SPR entailments using a classifier trained on SPR data, we noticeably outperform the majority baseline but are below USEF. 
Both ours and USEF's accuracies are lower than \newcite{teichert2017semantic}'s best reported numbers. This is not surprising as \citeauthor{teichert2017semantic} condition on observed semantic role labels when predicting proto-role labels.

\tabref{tab:spr_breakdown} reports accuracies for each proto-role. Whenever one of the classifiers outperforms the baseline for a proto-role, all the other classifiers do as well.
The classifiers outperform 
the majority baseline for $6$ of the reported $16$ proto-roles. 
We observe these $6$ properties are more associated with proto-agents than proto-patients. 

The larger improvements over the majority baseline for SPR compared to FN+ and DPR is not surprising. 
\newcite{dowty1991thematic} posited that proto-agent, and -patient should correlate
with English syntactic subject, and object, respectively, and
empirically the \emph{necessity of [syntactic] parsing for predicate
argument recognition} has been observed in practice~\cite{gildea2002necessity,punyakanok2008importance}. 
Further, recent work is suggestive that LSTM-based
frameworks implicitly may encode syntax based on certain learning
objectives~\cite{linzen2016assessing,shi-padhi-knight:2016:EMNLP2016,belinkov-EtAl:2017:I17-1}.
It is unclear whether NMT encoders capture semantic proto-roles specifically or just
underlying syntax that affects the proto-roles. 

\renewcommand{\d}{{\small $^\dagger$}}
\begin{table}
	\centering
    \begin{adjustbox}{width=1\columnwidth}
\begin{tabular}[h!]{l|cccc|c|hc}
\toprule 
\small{Proto-Role}	& ar	& es	& zh 	& de	& avg & \citeauthor{teichert2017semantic} & MAJ\\ \midrule 
\small{physically existed} & 70.6	& 70.8	& {\bf 77.2} 	& 70.8	& 72.4\d	&  & 65.9 \\
\small{sentient}	& 78.5	& {\bf 82.2}	& 80.5 	& 81.7	& 80.7\d	& 85.6 & 75.5 \\
\small{aware}	& 75.9	& {\bf 77.0}	& 76.6 	& 76.7	& 76.6\d	& 87.3 & 60.9 \\
\small{volitional} & 74.3	& {\bf 76.8}	& 74.7 	& 73.7	& 74.9\d	& &  64.5 \\
\small{existed before}	& 68.4	& {\bf 70.5}	& 66.5 	& 68.4	& 68.5\d	& &  64.8 \\
\small{caused}& 69.4	& {\bf 74.1}	& 72.2 	& 72.7	& 72.1\d	& &  63.4 \\ 
\midrule 
\small{changed} & 64.2	& 62.4	& 63.8 	& 62.0	& 63.1	 & & {\bf 65.1} \\
\small{location} & 91.1	& 90.1	& 90.4 	& 90.2	& 90.4	& & \textbf{91.7} \\
\small{moved} & 90.6	& 88.8	& 90.1 	& 90.3	& 89.9	&  & {\bf 93.3} \\
\small{used in} & 34.9	& 38.1	& 31.8 	& 34.2	& 34.7	&  & {\bf 55.2} \\
\small{existed after}	& 62.7	& 69.0	& 65.6 	& 65.2	& 65.7	&  & {\bf 69.7} \\
\small{chang. state}& 61.8	& 60.7	& 60.9 	& 60.7	& 61.0	& &  {\bf 65.2} \\
\small{chang. possession}& 89.6	& 88.6	& 89.9 	& 88.3	& 89.1	& &  {\bf 93.9} \\
\small{stationary during}	& 86.3	& 84.4	& 90.5 	& 86.0	& 86.8	& &  {\bf 96.3} \\	
\small{physical contact} & 85.0	& 82.0	& 84.5 	& 84.4	& 84.0	& &  {\bf 85.8} \\
\small{existed during}	& 59.3	& 71.8	& 60.8 	& 64.4	& 64.1	& &  {\bf 84.7} \\
\bottomrule 
\end{tabular}
\end{adjustbox}
\caption{\capsize Accuracies on the SPR test set 
broken down by each proto-role. ``avg'' represents the score for the proto-role averaged across 
target languages. Bold and \d \space respectively indicate the best results for each proto-role and whether all of our classifiers outperformed the proto-role's majority baseline. 
}
\label{tab:spr_breakdown}
\end{table}

\paragraph{NMT target language}
Our experiments show differences based on which target 
language was used to train the NMT encoder, in capturing semantic proto-roles and paraphrastic inference. In \tabref{tab:rte-results}, we notice a large improvement 
using sentence representations from an NMT encoder that was trained on en-es 
parallel text. The improvements are most profound when
a classifier trained on DPR data 
predicts entailment focused on 
semantic proto-roles 
or paraphrastic inference. We also note that using the NMT encoder trained on 
en-es 
parallel text results in the highest results in $5$ of the $6$ proto-roles in the top portion 
of \tabref{tab:spr_breakdown}.
When using other 
sentence representations (\appref{app:repr}), we notice that 
using representations from English-German encoders 
consistently outperforms 
using the other encoders~(Tables~\ref{tab:results-rte-infersent} and~\ref{tab:results-nli-infersent}).
This prevents us from making generalizations regarding specific target side languages.

\paragraph{NLI across multiple domains}
Though our main focus is exploring what NMT encoders learn about distinct semantic phenomena, we would like to know how useful NMT models are for general 
NLI across multiple domains. Therefore, we also evaluate the sentence representations with 
Multi-NLI. As indicated by \tabref{tab:nli-results}, the representations perform noticeably better than a majority baseline. 
However, 
our results are  
not competitive with state-of-the-art 
systems trained specifically for Multi-NLI~\cite{nangia2017repeval}.

\begin{table}[t]
	\centering
    \small
    \begin{tabular}{l|cccc|c}
    \toprule 
    	 & ar & es & zh & de & MAJ \\ \midrule 
    MNLI-1 & 45.9 & 45.7 & 46.6 & 48.0 & 35.6 \\
    MNLI-2 & 46.6 & 46.7& 48.2 & 48.9 & 36.5\\ 
    \bottomrule 
    \end{tabular}
    \caption{\capsize 
    Accuracies for MNLI test sets. MNLI-1 refers to the matched case and MNLI-2 is the mismatched.}
    \label{tab:nli-results}
\end{table}

\section{Related Work}
In concurrent work, 
\newcite{hypoths-nli} explore whether NLI datasets contain statistical irregularities by training a model with access to only hypotheses.  
Their model significantly outperforms the majority baseline and our results on Multi-NLI, SPR, and FN+.
They suggest that 
these, 
among other NLI datasets, contain statistical irregularities. 
Their findings illuminate
issues with the recast datasets we consider, 
but do not invalidate our approach of using recast NLI 
to determine whether NMT encoders capture distinct semantic phenomena. Instead, they force us to re-evaluate the majority baseline as an indicator of whether 
encoders learn distinct semantics
and to what extent we can make conclusions based on these recast datasets. 

Prior work has focused on the relationship between semantics and machine translation. 
MEANT
and 
its extension 
XMEANT 
evaluate MT systems based on
semantics~\cite{lo2011meant,lo2014xmeant}. Others have focused on incorporating semantics directly in MT. \newcite{P07-1005} use word sense disambiguation to help statistical MT, \newcite{W11-1012} add semantic-roles to improve phrase-based MT, and 
\newcite{W17-3209} demonstrate how filtering parallel sentences that are not parallel in meaning improves translation. 
Recent work explores how representations learned by NMT systems can improve semantic tasks. \newcite{NIPS2017_7209} show improvements in 
many tasks by using contextualized word vectors extracted from a LSTM encoder trained for MT. Their 
goal is to use NMT to improve other tasks while 
we focus on using NLI to determine 
what NMT models learn about different semantic phenomena.

Researchers have 
explored 
what NMT models learn about other linguistic 
phenomena,
such as  morphology~\cite{dalvi-EtAl:2017:I17-1,P17-1080}, syntax~\cite{shi-padhi-knight:2016:EMNLP2016}, and lexical semantics~\cite{belinkov-EtAl:2017:I17-1}, 
including
word senses~\cite{Marvin2018,liu2018handling}

\section{Conclusion and Future Work}
Researchers suggest that NMT models learn sentence representations that capture meaning.
We inspected whether distinct types of semantics are captured by NMT encoders. Our experiments 
suggest that NMT encoders might learn the most about semantic proto-roles, do not focus on anaphora resolution, and may poorly capture paraphrastic inference. 
We conclude by suggesting that target-side language affects how well an NMT 
encoder captures these semantic phenomena.

In future work, we would like to study how well NMT encoders capture other semantic phenomena,
possibly by recasting other datasets.
Comparing how semantic phenomena are represented in different NMT architectures,  
e.g. purely convolutional~\cite{pmlr-v70-gehring17a} or attention-based~\cite{NIPS2017_7181}, may 
shed light on whether 
different architectures may better capture  
semantic phenomena. 
Finally, investigating how multilingual systems learn semantics can bring a new perspective to questions of universality of representation~\cite{schwenk2017learning}.

\section*{Acknowledgements}
This work was supported by 
JHU-HLTCOE,
DARPA LORELEI, and Qatar Computing Research Institute. 
We  
thank
anonymous reviewers. 
Views and conclusions contained in this publication
are those of the authors and should not be
interpreted as representing official policies or endorsements
of DARPA or the U.S. Government.

\bibliography{references-no-url}
\bibliographystyle{acl_natbib}

\appendix
\section{Sentence Representations} 
\label{app:repr}
In the experiments reported in the main paper, we used a simple sentence representation, the first and last hidden states of the forward and backward encoders. We concatenated them for both the context and the hypothesis  and fed to a linear classifier. Here we compare the results of \texttt{InferSent}~\cite{D17-1070}, a more involved representation 
that was found to provide a good  
sentence representation based on NLI data.    
Specifically, we concatenate the forward and backward encodings for each sentence, and max-pool over the length of the sentence, resulting in $\hVect$ and $\tVect$ (in $ \reals^{2d}$) for the \context~and \hypothesis. The \texttt{InferSent} representation is defined by 
\begin{equation*}
(\tVect, \hVect, |\tVect - \hVect |, \tVect * \hVect ) \in \reals^{8d}
\end{equation*}
where the product and subtraction are carried element-wise and commas denote vector-concatenation. 

The pair representation is fed into a multi-layered perceptron (MLP) with one hidden layer and a ReLU non-linearity. We set the hidden layer size to 500 dimensions, similarly to~\newcite{D17-1070}. The softmax layer maps onto the number of labels, which is either 2 or 3 depending on the dataset. 

\paragraph{\texttt{InferSent} results}
\tabref{tab:results-rte-infersent} shows the results of the classifier trained on NMT representations with the InferSent architecture. Here, the representations from 
NMT encoders trained on the English-German
parallel corpus slightly outperforms the others. Since this data used a different corpus compared to the other language pairs, we cannot determine whether the improved results are due to the different target side language or corpus.
The main difference with respects to the simpler sentence representation (Concat) is improved results on FN+. \tabref{tab:results-nli-infersent} shows the results on Multi-NLI. 
It is interesting to note that, when using the sentence representations from NMT encoders, concatenating the sentence vectors outperformed the \texttt{InferSent} method on Multi-NLI.

\begin{table}[t]
\centering
\small
\begin{minipage}[t]{\linewidth}
\centering
\begin{tabular}{p{2cm}l|lll}
\toprule
& & FN+ & DPR & SPRL  \\
\midrule
\multirow{4}{2cm}{NMT Concat} & en-ar & 56.2 & 49.8 & 72.1 \\
&  en-es & 56.1 & 50.0 & 74.2 \\ 
&  en-zh & 54.3 & 50.0 & 73.1 \\
&  en-de & 55.5 & 50.0 & 73.1 \\
\midrule
\multirow{4}{2cm}{NMT \texttt{InferSent}} & en-ar & 57.9 & 50.0 & 73.6 \\
 &  en-es & 58.0 & 50.0 & 72.7 \\ 
 &  en-zh & 57.8 & 49.8 & 72.4 \\
 &  en-de & 58.3 & 50.1 & 73.7 \\
\midrule
\multicolumn{2}{l|}{Majority} & 57.5 & 50.0 & 65.4  \\
\multicolumn{2}{l|}{\cite{white-EtAl:2017:I17-1}} & 80.5 & 49.5 & 80.6 \\ 
\bottomrule
\end{tabular}
\caption{\capsize NLI results on fine-grained semantic phenomena. FN+ = paraphrases; DPR = pronoun resolution; SPRL = proto-roles. NMT representations are combined with either a simple concatenation (results copied from Table~\ref{tab:rte-datasets}) or the \texttt{InferSent} representation. State-of-the-art (SOTA) is from \newcite{white-EtAl:2017:I17-1}.}
\label{tab:results-rte-infersent}
\end{minipage} \\ 
\vspace{2cm}
\begin{minipage}[t]{\linewidth}
\centering
\small
\begin{tabular}{p{1.5cm}l|hll}
\toprule
& & SNLI & MNLI-1 & MNLI-2  \\
\midrule
\multirow{4}{1cm}{NMT Concat} & en-ar &  & 45.9 & 46.6 \\
&  en-es &  & 45.7 & 46.7\\ 
&  en-zh &  & 46.6 & 48.2 \\
&  en-de &  & 48.0 & 48.9 \\
\midrule
\multirow{4}{1cm}{NMT \texttt{Infer}-\texttt{Sent}} & en-ar & \hl{44.19} & 40.1 & 41.8 \\
&  en-es & \hl{43.94} & 44.9 & 40.8 \\ 
&  en-zh & \hl{44.34} & 43.7 & 42.1\\
&  en-de & \hl{44.84} & 41.3 & 41.1\\
\midrule
\multicolumn{2}{l|}{Majority} & & 35.6 & 36.5  \\
\multicolumn{2}{l|}{SOTA} & & 81.10 & 83.21 \\ 
\bottomrule
\end{tabular}
\caption{\capsize Results on language inference 
on MultiNLI~\cite{williams2017broad}, matched/mismatched scenario (MNLI1/2).}
\label{tab:results-nli-infersent}
\end{minipage}
\end{table}

\section{Implementation \& Experimental Details}
\label{app:exp-details}
 We use $4$-layer NMT systems with 500-dimensional word embeddings and LSTM states (i.e., $d=500$). The vocabulary size is $75$K words. We train NMT models until convergence and take the models that performed best on the development set for generating representations to feed into the entailment classifier.  
We use the hidden states from the top encoding layer for obtaining sentence representations
since it has been hypothesized that higher layers focus on word meaning, as opposed to syntax~\cite{P17-1080,belinkov-EtAl:2017:I17-1}.
We remove long sentences ($>50$ words) when training both the classifier and the NMT model, as  is common NMT practice~\cite{cho-EtAl:2014:SSST-8}. 
During testing, we use all test sentences regardless of sentence length. Our implementation extends \newcite{P17-1080}'s implementation in Torch~\cite{collobert2011torch7}. 

We train English$\rightarrow$Arabic/Spanish/Chinese NMT models on the first 2 million sentences of the United Nations parallel corpus training set~\cite{ZIEMSKI16.1195}, and the English$\rightarrow$German model on the WMT dataset~\cite{bojar-EtAl:2014:W14-33}. We use the official training/development/test splits. 

In our NLI experiments, we do not train on Multi-NLI and test on the recast datasets, or vice-versa, since 
Multi-NLI since Multi-NLI uses a 3-way classification (\textit{entailment}, \textit{neutral}, and \textit{contradictions}) while the recast datasets use just two labels (\textit{entailed} and \textit{not-entailed}).
In preliminary experiments, we also used a 3-layered MLP. Although the results slightly improved, we noted similar trends to the linear classifier.

\section{Sentence length}
\label{app:data-diff}
The average sentence in the FN+ test dataset is $31$ words and almost $10\%$ of the test sentences are longer than $50$ words. In SPR and DPR, each premise sentence has on average $21$ and $15$ words respectively and only $1\%$ of sentences in SPR have more than $50$ words. No DPR sentences have $>50$ words.

\tabref{tab:fnplus-sent-len} reports accuracies for ranges of sentence lengths in FN+'s development set. When trained on sentence representations form an English$\rightarrow$Chinese,German NMT encoder, the NLI accuracies steadily decrease. When using English$\rightarrow$Arabic, the accuracies stay consistent until sentences have between $70$--$80$ tokens while the results from English$\rightarrow$Spanish quickly drops from $0$--$10$ to $10$--$20$ but then stays relatively consistent.

\section{World Knowledge in DPR}
\label{sec:wk-dpr}
When released, \newcite{rahman-ng:2012:EMNLP-CoNLL}'s DPR dataset confounded the best co-reference models because ``its difficulty stems in part from its reliance on sophisticated knowledge sources.'' ~\tabref{tab:dpr-example} includes examples that demonstrate how world knowledge is needed to accurately predict these recast NLI sentence-pairs. 

\textcolor{white}{\lipsum[1]}
\begin{table}[t!]
\begin{adjustbox}{width=1\linewidth}
\centering
\begin{tabular}{c|cccc|r}
\toprule
\small{Sentence length} & ar & es & zh & de & total\\
\midrule
\small{0-10} &  \small{46.8} &  \small{63.7} &  \small{66.0} &  \small{65.4} & \small{526} \\ 
\small{10-20} &  \small{49.0} &  \small{53.3} &  \small{57.4} &  \small{56.5} & \small{2739} \\ 
\small{20-30} &  \small{48.4} &  \small{54.0} &  \small{53.2} &  \small{54.9} & \small{4889} \\ 
\small{30-40} &  \small{48.4} &  \small{54.1} &  \small{51.2} &  \small{53.9} & \small{4057} \\ 
\small{40-50} &  \small{47.7} &  \small{59.0} &  \small{55.0} &  \small{58.7} & \small{2064} \\ 
\small{50-60} &  \small{49.1} &  \small{56.1} &  \small{54.5} &  \small{57.5} & \small{877} \\ 
\small{60-70} &  \small{46.4} &  \small{53.6} &  \small{43.9} &  \small{44.1} & \small{444} \\ 
\small{70-80} &  \small{59.9} &  \small{51.6} &  \small{43.3} &  \small{43.3} & \small{252} \\ 
\bottomrule
\end{tabular}
\end{adjustbox}
\caption{\capsize Accuracies on FN+'s dev set based on sentence length. The first column represents the range of sentences length: first number is inclusive and second is exclusive. The last column represents how many context sentences have lengths that are in the given row's range.}
\label{tab:fnplus-sent-len}
\end{table}

\begin{table*}[t!]
\vspace{7.5pt}
\begin{adjustbox}{width=1\linewidth}
\centering
\begin{tabular}{l|c}
\toprule
Chris was running after John, because he stole his watch & \\
$\blacktriangleright$ Chris was running after John, because John stole his watch & \cmark \\
$\blacktriangleright$ Chris was running after John, because Chris stole his watch & \xmark \\  \midrule
Chris was running after John, because he wanted to talk to him & \\
$\blacktriangleright$ Chris was running after John, because Chris wanted to talk to him  & \cmark \\
$\blacktriangleright$ Chris was running after John, because John wanted to talk to him & \xmark \\ \midrule
The plane shot the rocket at the target, then it hit the target & \\
$\blacktriangleright$ The plane shot the rocket at the target, then the rocket hit the target & \cmark \\
$\blacktriangleright$ The plane shot the rocket at the target, then the target hit the target& \xmark \\ \midrule 
Professors do a lot for students, but they are rarely thankful & \\
$\blacktriangleright$ Professors do a lot for students, but students are rarely thankful & \cmark \\
$\blacktriangleright$ Professors do a lot for students, but Professors are rarely thankful & \xmark \\ \midrule
MIT accepted the students, because they had good grades & \\
$\blacktriangleright$ MIT accepted the students, because the students had good grades & \cmark \\ 
$\blacktriangleright$ MIT accepted the students, because MIT had good grades & \xmark \\ \midrule
Obama beat John McCain, because he was the better candidate & \\
$\blacktriangleright$ Obama beat John McCain, because Obama was the better candidate & \cmark \\ 
$\blacktriangleright$  Obama beat John McCain, because John McCain was the better candidate & \xmark \\ \midrule
 Obama beat John McCain, because he failed to win the majority of the\\ electoral votes & \\
 $\blacktriangleright$ Obama beat John McCain, because John McCain failed to win\\ \hspace{10pt} the majority of the electoral votes & \cmark \\ 
  $\blacktriangleright$ Obama beat John McCain, because Obama failed to win\\ \hspace{10pt} the majority of the electoral vote & \xmark \\ \bottomrule
\end{tabular}
\end{adjustbox}
\caption{\capsize Examples 
from DPR's dev set. The first line in each section is a context and 
lines with $\blacktriangleright$ are corresponding hypotheses. \cmark \space (\xmark) in the last column indicates whether the hypothesis is entailed (or not) by the context.}
\label{tab:dpr-example}
\end{table*}

\end{document}